\documentclass{article} 
\usepackage{iclr2027_conference,times}

\usepackage{hyperref}
\usepackage{url}

\usepackage{amsmath}
\usepackage{amssymb}
\usepackage{graphicx}
\usepackage{booktabs}
\usepackage{multirow}
\usepackage{colortbl}
\usepackage{algorithm}
\usepackage{algorithmic}

\definecolor{linearKVGray}{gray}{0.93}
\definecolor{macroHL}{rgb}{0.83,0.90,0.98}

\title{LinearKV: One Cached State Suffices for \\ Position-Independent Caching in Hybrid LLMs}

\author{
Yirui Liu$^{1,*}$\quad Ruoling Qi$^{2,1,*}$\quad Longwen Wang$^{3,1}$\quad Xuaner Wu$^{1}$\quad Jian Chen$^{4}$ \\
\bfseries Yuxin Jin$^{1}$\quad Jiawei Shao$^{1}$\quad Xuelong Li$^{1,\dagger}$ \\[5pt]
{\normalfont\small $^{1}$Institute of Artificial Intelligence, China Telecom (TeleAI)} \\
{\normalfont\small $^{2}$Shanghai Jiao Tong University \quad $^{3}$Xi'an Jiaotong University \quad $^{4}$University at Buffalo} \\[3pt]
{\normalfont\small $^{*}$Equal contribution.\quad $^{\dagger}$Corresponding author.} \\[3pt]
{\normalfont\small Yirui Liu: \texttt{yiruiliu926@gmail.com}\quad Ruoling Qi: \texttt{qiruoling760@sjtu.edu.cn}} \\
{\normalfont\small Xuelong Li: \texttt{xuelong\_li@ieee.org}}
}

\iclrfinalcopy
\begin{document}

\maketitle

\lhead{}
\renewcommand{\headrulewidth}{0pt}

\begin{abstract}
LLM serving is increasingly accelerated by position-independent caching (PIC). Existing PIC methods, however, are built for full-attention models, where a token-indexed KV cache underlies its core operations: matching reusable token chunks, concatenating their KV entries, and selectively recomputing a few tokens to restore cross-chunk context. Hybrid LLMs break these primitives---they replace most attention layers with linear recurrences that expose only a fixed-size state, leaving no token-indexed KV to concatenate or to locally repair. This raises a natural question: can PIC benefit hybrid models, and what would it take? We present LinearKV, a training-free hybrid-PIC framework. Its key insight is a \emph{decoupled initialization}: each linear layer maps its $K$ matched local states to a single initial state, while full-attention layers concatenate their KV as before. LinearKV is therefore compatible with existing PIC methods, reusing their token selection and recomputation as-is. Under this framework, we find that a \emph{single cached state} suffices as the linear layer's initializer. The algebraically principled alternative---composing all $K$ cached states into the exact full-prefix state, as concurrent work HYPIC does---is unnecessary and, on some architectures, even harmful. We compare the two across three hybrid models (one Mamba-2, two GDN) and three PIC selectors (CacheBlend, EPIC, ProphetKV). On the two GDN models the two tie, both recovering most of full quality (up to $92\%$); on the Mamba-2 model, exact composition instead collapses under every selector---under EPIC, for instance, it recovers only $46.6\%$ of full quality, versus $86.8\%$ for a single cached block initializer. A single state initializer is also cheaper, cutting time-to-first-token to $0.46\times$ full prefill versus a further $5$--$17\%$ overhead for exact composition; results hold across LongBench QA and RULER at 8K--32K.
\end{abstract}

\section{Introduction}
\label{sec:intro}

Long-context LLM applications---multi-turn conversations, agentic workflows, long-document understanding---incur a prefill cost that grows with prompt length, inflating time to first token (TTFT) and reducing throughput. Context (or prompt) caching mitigates this by storing KV-cache entries for previously processed tokens and reusing them across requests. Standard implementations are prefix-based \citep{kwon2023vllm,zheng2024sglang}: a cached entry is reusable only when the same tokens follow the same prefix, limiting the hit rate.

Position-independent caching (PIC) relaxes this constraint, letting an independently cached chunk be reused at any position. Reuse then becomes lossy---a chunk prefilled in isolation lacks the cross-chunk attention a joint prefill would supply---so PIC methods selectively recompute a small subset of tokens to restore the cross-chunk attention \citep{yao2025cacheblend,hu2024epic,yang2025kvshare,prophetkv2026}. Offline, each chunk is prefilled independently and its KV stored under an identifier; online, a request retrieves the matching chunks, concatenates their KV in context order, recomputes selected positions, and serves from the repaired cache.

\begin{figure}[t]
\centering
\includegraphics[width=0.50\textwidth]{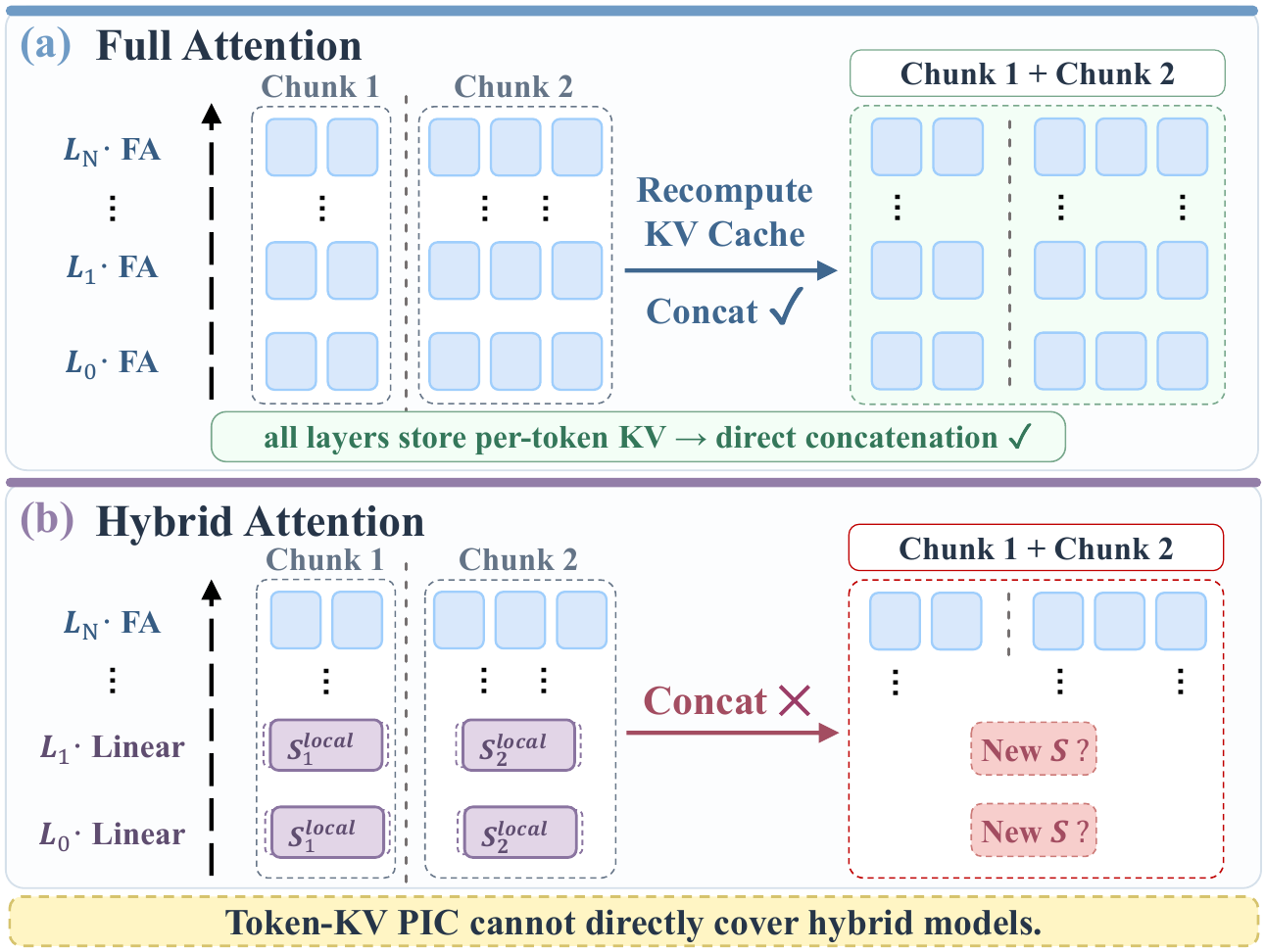}
\caption{Why token-KV PIC does not transfer directly to hybrid models. \textbf{(a)}~Full attention: every layer stores per-token KV that concatenates across chunks. \textbf{(b)}~Hybrid: recurrent layers expose only a per-chunk state, which does not.}
\label{fig:intro}
\end{figure}

The PIC methods above are designed for full-attention models. Hybrid architectures, however, have emerged as a promising long-context backbone: they interleave a few full-attention (FA) layers with a majority of recurrent layers such as Mamba-2 \citep{dao2024ssd} and Gated DeltaNet (GDN) \citep{yang2025gateddeltanet}. This raises a natural question: Can PIC benefit hybrid models, and what would it take? Answering it exposes two challenges. \textbf{(C1) Can the existing PIC machinery be reused, or must hybrid PIC be redesigned?} The reuse step is where the mismatch surfaces (Figure~\ref{fig:intro}): FA layers still expose a per-token KV cache to concatenate, but each recurrent layer summarizes its whole chunk into a single fixed-size state, leaving no per-token entries to concatenate or selectively recompute. Chunk matching still works by identifier, but PIC's other core operations---concatenation and selective repair---assume a token-indexed cache the recurrent path lacks, so whether PIC can work on this stateful representation at all is unclear. \textbf{(C2) Does hybrid PIC generalize across models?} Hybrids vary in recurrence family (Mamba-2 vs.\ GDN) and recurrent-to-FA ratio, so whether one construction stays beneficial across these designs---or must be re-tuned per architecture---is unknown.

We present LinearKV, a training-free PIC framework for hybrid models. Its key insight is a \emph{decoupled initialization} applied right after token-chunk matching: each FA layer is initialized by concatenating its cached KV entries, exactly as in full-attention PIC, while each linear layer is initialized by a function that maps the $K$ matched local states to a single initial state. As it leaves the FA path and the selector interface untouched, this decoupling lets LinearKV reuse existing PIC methods as a plug-in for choosing which tokens to recompute (C1): LinearKV then applies that choice across both layer types---overwriting the selected FA KV while updating each linear state through the selected positions in context order. Moreover, LinearKV shows that \emph{one cached state suffices} as the linear-layer initializer: keeping just a single matched chunk's cached linear state is remarkably robust across architectures (C2). It matches the algebraically principled alternative---composing all $K$ cached states into the exact full-prefix state, as concurrent HYPIC \citep{liu2026hypic} does---on the two GDN models, while on Mamba-2, exactly where that exact composition collapses, a single cached state initializer lifts quality from $46.6\%$ to $86.8\%$ of full. A \emph{random} single block works as well (Appendix~\ref{app:singlesource}), so it is the single-source construction, not the specific block, that matters.

We evaluate LinearKV on three hybrid models---Granite (Mamba-2), and OLMo and Qwen (GDN)---covering both recurrence families and different linear-to-FA ratios. On each, CacheBlend, EPIC, and ProphetKV serve as three token-recomputation selectors, and we compare the last-block initializer against exact composition under identical selection and recomputation. Across LongBench QA and RULER at 8K--32K, the same architecture-dependent boundary holds---exact composition collapses on Mamba-2 under every selector, while the last-block initializer stays robust and matches it on the two GDN models---and LinearKV also reduces TTFT relative to full prefill in all evaluated settings.

In summary, this paper makes three contributions:
\begin{itemize}
\item We formalize position-independent caching for hybrid LLMs as a decoupled-initialization framework, identifying \emph{linear-state initialization}---mapping the $K$ cached states to one initial state---as the single hybrid-specific operation, within which existing full-attention PIC selectors are reused unchanged.
\item We show that the natural choice---composing all $K$ cached states (exact composition, as in concurrent HYPIC)---is unnecessary and architecture-fragile: \emph{one cached state suffices}. A single-block initializer with ordered recomputation matches exact composition on GDN, lifts Mamba-2 from $46.6\%$ to $86.8\%$ of full quality, and avoids its per-request composition cost; a random single block does as well (Appendix~\ref{app:singlesource}), isolating single-source construction, not the specific block, as the operative factor.
\item We evaluate LinearKV across three hybrid models and three PIC selectors on LongBench QA and RULER (8K--32K), where a training-free last-block initializer is robust across architectures and reduces TTFT relative to full prefill in all evaluated settings.
\end{itemize}

\section{Preliminaries}
\label{sec:preliminaries}

Hybrid LLMs interleave a small number of FA layers for global token--token interaction with a majority of recurrent layers, such as Mamba-2 \citep{dao2024ssd} and Gated DeltaNet (GDN) \citep{yang2025gateddeltanet}, that summarize the processed prefix in fixed-size state.

An FA layer retains a token-indexed KV cache $\{(K_i,V_i)\}_{i=1}^{N}$. A recurrent layer instead compresses the processed prefix into a fixed-size state $S_i^\ell\in\mathbb{R}^{d_k\times d_v}$, updated token by token through the general affine recurrence
\begin{equation}
S_i^\ell = T_i^\ell S_{i-1}^\ell + u_i^\ell,
\label{eq:recurrence}
\end{equation}
where $T_i^\ell$ transports the previous state and $u_i^\ell$ is a token-dependent outer-product update. The two recurrence families instantiate $T_i^\ell$ differently. Suppressing the layer superscript for clarity,
\begin{align}
\text{Mamba-2:}\quad
&S_i = a_i S_{i-1}+u_i,
&&T_i=a_i I, \label{eq:mamba-update}\\
\text{GDN:}\quad
&S_i = T_iS_{i-1}+u_i,
&&T_i=\alpha_i(I-\beta_i k_i k_i^\top), \label{eq:gdn-update}
\end{align}
where $a_i\in(0,1]$ is a per-head scalar decay, $\alpha_i=\exp(g_i)$ is a learned gate, and $I-\beta_i k_i k_i^\top$ is a token-dependent rank-1 correction. Mamba-2 therefore transports every state direction within a head using the same scalar, whereas GDN applies a direction-dependent dense transition. This difference later helps interpret the depth-wise reuse error analyzed below. After prefill, only the final state is required for subsequent decoding; the intermediate states do not form a persistent token-indexed cache.

\section{Method: LinearKV}
\label{sec:method}

We propose LinearKV, a training-free PIC framework for hybrid LLMs such as Granite, OLMo, and Qwen. Its key idea is a \emph{decoupled linear-state initialization}: each full-attention (FA) layer is initialized by concatenating its cached KV entries, exactly as in full-attention PIC, while each linear layer is initialized by a function $f$ that maps the $K$ matched local states to a single initial state. Because this decoupling leaves the FA path and the selector interface untouched, LinearKV requires only that a PIC selector emit a set of token positions to recompute; any selector meeting that interface plugs in unchanged---LinearKV reuses its token selection and recomputation as-is---and the design space collapses to the one remaining choice, the linear-state initializer $f$. We validate this with three selectors (CacheBlend, EPIC, ProphetKV). This section first gives an overview of LinearKV, then studies the choice of $f$ in depth, showing that the algebraically exact composition fails on some architectures while a simple last-block initialization is robust across models and selectors---recovering most of full quality on the GDN models and, on Mamba-2, far exceeding exact composition.

\subsection{Overview}

\begin{figure}[t]
\centering
\includegraphics[width=\textwidth]{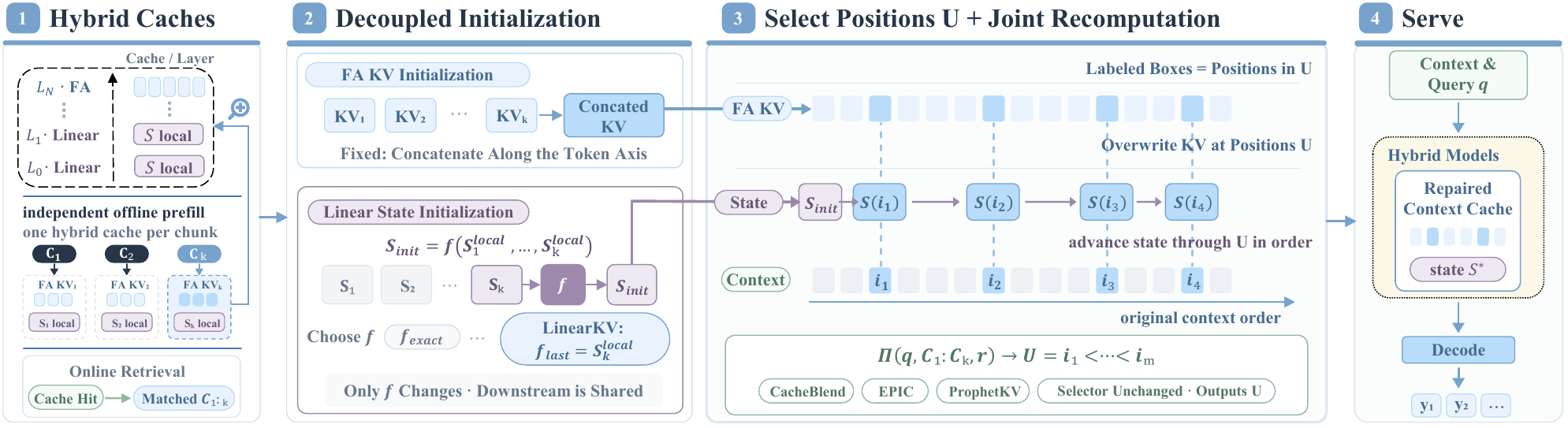}
\caption{Overview of LinearKV. \textbf{(1)}~Each reusable token chunk $C$ is prefilled offline into one hybrid cache---per-chunk FA KV together with, for every recurrent layer, an end state $S^{\mathrm{local}}$; online retrieval returns the matched chunks $C_{1:K}$. \textbf{(2)}~\emph{Decoupled initialization}: FA layers concatenate their cached KV along the token axis (fixed), while each recurrent layer folds the $K$ cached states into one initial state $S_{\mathrm{init}}=f(S^{\mathrm{local}}_1,\dots,S^{\mathrm{local}}_K)$. Only $f$ changes---exact composition versus LinearKV's last block $f_{\mathrm{last}}=S^{\mathrm{local}}_K$---and everything downstream is shared. \textbf{(3)}~Any existing selector $\Pi(q,C_{1:K},r)$ outputs repair positions $\mathcal{U}=\{i_1<\dots<i_m\}$; LinearKV replays $\mathcal{U}$ in original context order, advancing each recurrent state from $S_{\mathrm{init}}$ while overwriting the FA KV at those positions. \textbf{(4)}~The query is served from the repaired context cache.}
\label{fig:overview}
\end{figure}

Figure~\ref{fig:overview} summarizes the pipeline. \emph{Offline}, each reusable chunk $C$ is prefilled independently and cached as its per-chunk FA KV together with each recurrent layer's local state $S^{\mathrm{local}}$ (the last-block initializer needs nothing more; exact composition additionally uses each chunk's transition, obtained as described in \emph{Linear-State Initialization} below). \emph{Online}, given the $K$ matched chunks, the FA path proceeds as in full-attention PIC, while each recurrent layer forms its initial state through the initializer $f$:
\begin{equation}
S_{\mathrm{init}} = f\!\left(S^{\mathrm{local}}_1, \dots, S^{\mathrm{local}}_K\right).
\label{eq:init-fn}
\end{equation}
LinearKV then adopts existing PIC method---CacheBlend, EPIC, or ProphetKV---to select a token position set $\mathcal{U}=\{i_1<\dots<i_m\}$ for recomputation, with $m\approx rN$ under budget $r$, and repairs the cache by processing \emph{only} these $m$ tokens in context order (Algorithm~\ref{alg:linearkv}). The intervening unselected tokens are never re-run: each FA layer recomputes the selected tokens against the assembled cache and overwrites their KV, while each recurrent layer advances its state from $S_{\mathrm{init}}$ through the selected tokens alone (Eq.~\ref{eq:recurrence}), keeping the final state. Dropping the unselected transitions makes the recomputation itself an approximation---one whose cost is $m\approx rN$ token-steps rather than a full prefill---after which the query is prefilled against the repaired cache and served.

\begin{algorithm}[t]
\caption{LinearKV serving of one request}
\label{alg:linearkv}
\begin{algorithmic}[1]
\REQUIRE matched chunks $C_{1:K}$ (cached FA KV + recurrent end states $S^{\mathrm{local}}_j$; exact composition also uses transitions $T_{C_j}$); query $q$; selector $\Pi$; budget $r$
\STATE \textbf{Init (decoupled):} FA layers concatenate cached KV in context order; each recurrent layer sets $S_{\mathrm{init}}{\leftarrow}f(S^{\mathrm{local}}_1,\dots,S^{\mathrm{local}}_K)$ ($f{=}$ last block or exact)
\STATE $\mathcal{U}{\leftarrow}\Pi(q,C_{1:K},r){=}\{i_1{<}\dots{<}i_m\}$, $m{\approx}rN$ \COMMENT{selector unchanged}
\STATE \textbf{Recompute:} over the $m$ selected tokens $\mathcal{U}$ only, in context order, per layer---FA recomputes against the assembled cache and overwrites KV at $\mathcal{U}$; each recurrent layer advances $S$ from $S_{\mathrm{init}}$ through $\mathcal{U}$ (Eq.~\ref{eq:recurrence}) and keeps the final $S$
\STATE prefill $q$ against the repaired cache and decode
\end{algorithmic}
\end{algorithm}

\subsection{Linear-State Initialization}
\label{sec:exact}

The decoupled design of Eq.~\eqref{eq:init-fn} leaves one function unspecified: how each recurrent layer maps its $K$ cached local states $\{S^{\mathrm{local}}_j\}$ into a single initial state $S_{\mathrm{init}}$. This subsection compares two choices, and finds---counterintuitively---that the algebraically principled one is the fragile one.

\paragraph{Exact composition.} The algebra of the recurrence points to a mathematically intuitive first choice. Because the recurrence (Eq.~\eqref{eq:recurrence}) is linear in the state, running it through a whole chunk is itself an affine operator: the chunk takes whatever state $S_{\mathrm{in}}$ enters it, scales it by a cumulative transition $T_{C_j}=\prod_{t\in C_j}T_t$ (the product of the per-token transitions), and adds its own contribution $S^{\mathrm{local}}_j$---the state the chunk reaches from an empty start, which is exactly what we cache offline:
\begin{equation}
S_{\mathrm{out}} = T_{C_j}\,S_{\mathrm{in}} + S^{\mathrm{local}}_j .
\label{eq:chunk-map}
\end{equation}
To assemble the $K$ matched chunks, we apply these operators back to back in context order---chunk~1's output is chunk~2's input, and so on. This telescopes into a single closed form that composes the cached chunks with no approximation \emph{of its own}:
\begin{equation}
S_{\mathrm{init}}^{\mathrm{exact}}=\sum_{j=1}^{K}\Big(\prod_{m>j} T_{C_m}\Big)\,S^{\mathrm{local}}_j ,
\label{eq:exact-compose}
\end{equation}
each chunk's contribution carried forward by the transitions of all later chunks. This is the natural, algebraically principled choice, and the one concurrent HYPIC \citep{liu2026hypic} adopts. The two families differ only in what the transition $T_{C_m}$ is: Mamba-2's per-token transition is a scalar (Eq.~\eqref{eq:mamba-update}), so $T_{C_m}$ is a single scalar decay, whereas GDN's is a dense rank-1 update (Eq.~\eqref{eq:gdn-update}), so $T_{C_m}$ is a dense $d_k\times d_k$ matrix.

\paragraph{Exactness is conditional.} The exactness, though, is only relative to the chunk operators it composes---and each such operator, built offline from a chunk prefilled \emph{in isolation}, is not the operator that chunk would have inside a full prefill. Because an isolated chunk sees none of its predecessors, its operator is conditioned on inputs that omit all earlier context. This is harmless at the first layer, where every chunk starts from the same empty history, but not above it: each layer reads the previous layer's outputs, which in a full prefill already carry the earlier chunks' context. Composing exact algebra over these mis-conditioned operators therefore need not reproduce the true state. Concretely, let $e_j$ be the gap between the composed state and the true full-prefix state after chunk $j$; it follows a simple recursion,
\begin{equation}
e_j = T_{C_j}\,e_{j-1} + \delta_j ,
\label{eq:mismatch-propagation}
\end{equation}
where the first term carries the accumulated error forward and $\delta_j$ is the fresh mismatch injected because a chunk's isolated operator differs from its full-prefill operator. Whether the accumulated error is retained or damped as it is carried forward is modulated by the transition $T_{C_j}$---a scalar decay can only retain and sum it, whereas a dense gated transition can also suppress or overwrite particular directions. The injected term $\delta_j$ itself is not set by $T_{C_j}$: it also depends on the layer's hidden inputs, its depth and placement relative to the FA layers, and normalization. So the recurrence family shapes error \emph{propagation} but is not the sole cause of the outcome; whether exact composition actually helps is therefore an empirical question, which we settle next.

\begin{figure}[t]
\centering
\includegraphics[width=0.62\textwidth]{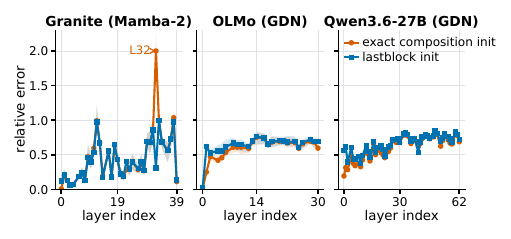}
\caption{Per-layer relative error of the constructed initial state vs.\ the true full-prefix state, at construction time (before recomputation), mean over five LongBench QA datasets. }
\label{fig:prerelerr}
\end{figure}

\paragraph{Diagnosis via relative error.} We answer it by probing the constructed initial state directly, measuring its per-layer relative error against the true full-prefix state $S^*$---i.e.\ $\lVert S_{\mathrm{init}}^\ell - S^{*\ell}\rVert / \lVert S^{*\ell}\rVert$ at each recurrent layer $\ell$, before any recomputation (Figure~\ref{fig:prerelerr}). At layer~0 the error is ${\sim}0.01$ on every model---near-exact, at the \texttt{bf16}/convolution-boundary rounding floor rather than truly zero---confirming that composition is faithful where the operators are correct (a numerical sanity check that the divergence at depth is model behavior, not an implementation artifact). At depth the two recurrence families diverge. On the GDN models (OLMo, Qwen) the error stays bounded ($<1$) at every layer. On Granite (Mamba-2) it compounds: one deep layer (L32) spikes to ${\sim}2\times$ the state norm---stable across all five benchmarks (range $1.7$--$2.3$)---with several other layers above $1.0$. This is exactly the split Eq.~\eqref{eq:mismatch-propagation} allows: Mamba-2's scalar decay retains and sums transported mismatch without a direction-dependent correction, whereas GDN's gate and dense rank-1 update can suppress or overwrite particular state directions. Neither the number nor the fraction of recurrent layers explains the split: Qwen has \emph{more} recurrent layers than Granite in absolute terms ($48$ GDN vs.\ $36$ Mamba-2), and Granite has the \emph{highest} recurrent fraction of the three ($90\%$ vs.\ $75\%$), yet Granite is the one that compounds---so within these models the governing factor tracks the recurrence dynamics, not how many layers are recurrent. We observe this on one Mamba-2 and two GDN hybrids; we therefore report it as an empirical boundary over the evaluated models, consistent with the error recursion of Eq.~\eqref{eq:mismatch-propagation}, rather than a proven law for every hybrid, and Eq.~\eqref{eq:mismatch-propagation} makes the prediction testable as more hybrids of each family appear.

\paragraph{Single-block initialization.} The diagnosis points to a fix. The failure is not inaccurate algebra but the aggregation of $K$ independently conditioned deep operators; a construction that reads from a \emph{single} cached block avoids it entirely. It discards $K{-}1$ of the states and keeps just one as the initial state, $S_{\mathrm{init}}=S^{\mathrm{local}}_{b}$ for some block $b$; by default we take the last matched chunk ($b{=}K$)---the most recent context, already cached and needing no extra selection---and \emph{which} block is kept turns out to barely matter, a random one doing as well (see Appendix~\ref{app:singlesource}). Such a single cached state throws away information exact composition preserves and is provably not the full-prefix state, so on its face it should be worse. The relative-error curves say otherwise (Figure~\ref{fig:prerelerr}, blue): the single block tracks exact composition on the GDN models---exactness buys nothing there---and on Granite it stays near $0.3$ exactly where exact composition explodes, with no spike at any depth. The algebraically principled construction is the fragile one; a single cached state is the robust one. Relative error, though, is only a construction-time proxy---it scores the \emph{starting point} recomputation is handed, not the served answer, and being norm-normalized it can be inflated where the target state norm is small. What ultimately matters is that this better starting point carries through to end-to-end quality once the selector's recomputation runs. That end-to-end quality is not norm-sensitive, and it moves the same way: the next section shows the Granite collapse and the GDN tie reappear in served accuracy across all three models and all three selectors (Tables~\ref{tab:main} and~\ref{tab:ruler}), confirming the proxy is tracking a real effect rather than a normalization artifact.

\section{Experiments}
\label{sec:experiments}

\begin{table}[t]
\centering
\caption{LongBench results under a matched ${\sim}20\%$ recompute budget ($r{=}0.2$): QA token-F1 per dataset and summarization ROUGE-L per dataset, each with its Avg highlighted. \textsc{ex}\,$=$ exact composition, \textsc{lb}\,$=$ LinearKV's last-block initial state (shaded rows). Summarization uses 512-token generation and the LongBench-official \texttt{rouge} implementation; na\"ive reuse already reaches $74$--$98\%$ of full there, so the spread between methods is compressed. $^{\ddagger}$Qwen anchors in the QA columns only are from an earlier torch build (method-independent). RULER is in Table~\ref{tab:ruler}.}
\label{tab:main}
\scriptsize
\setlength{\tabcolsep}{4.2pt}
\begin{tabular}{lccccc>{\columncolor{macroHL}}cccc>{\columncolor{macroHL}}c}
\toprule
& \multicolumn{6}{c}{LongBench QA (token-F1)} & \multicolumn{4}{c}{LongBench summarization (ROUGE-L)} \\
\cmidrule(lr){2-7} \cmidrule(lr){8-11}
Method & HQA & 2Wiki & MSQ & NQA & Qsp & Avg & QMSum & GovRep & MultiNews & Avg \\
\midrule
\multicolumn{11}{l}{\emph{Granite-4.0-H-Tiny (Mamba-2)}} \\
Full recompute & 0.386 & 0.351 & 0.210 & 0.186 & 0.421 & \cellcolor{macroHL} 0.311 & 0.237 & 0.325 & 0.265 & \cellcolor{macroHL} 0.276 \\
Na\"ive reuse & 0.204 & 0.149 & 0.103 & 0.112 & 0.190 & \cellcolor{macroHL} 0.152 & 0.203 & 0.196 & 0.211 & \cellcolor{macroHL} 0.204 \\
CacheBlend + \textsc{ex} & 0.043 & 0.085 & 0.024 & 0.025 & 0.140 & \cellcolor{macroHL} 0.063 & 0.186 & 0.090 & 0.161 & \cellcolor{macroHL} 0.146 \\
\rowcolor{linearKVGray} CacheBlend + \textsc{lb} & 0.087 & 0.124 & 0.077 & 0.103 & 0.199 & \cellcolor{macroHL} 0.118 & 0.209 & 0.283 & 0.243 & \cellcolor{macroHL} 0.245 \\
EPIC + \textsc{ex} & 0.194 & 0.180 & 0.083 & 0.060 & 0.205 & \cellcolor{macroHL} 0.145 & 0.200 & 0.162 & 0.229 & \cellcolor{macroHL} 0.197 \\
\rowcolor{linearKVGray} EPIC + \textsc{lb} & 0.353 & 0.242 & 0.222 & 0.212 & 0.320 & \cellcolor{macroHL} 0.270 & 0.242 & 0.302 & 0.263 & \cellcolor{macroHL} 0.269 \\
ProphetKV + \textsc{ex} & 0.128 & 0.168 & 0.042 & 0.059 & 0.226 & \cellcolor{macroHL} 0.125 & 0.209 & 0.180 & 0.206 & \cellcolor{macroHL} 0.198 \\
\rowcolor{linearKVGray} ProphetKV + \textsc{lb} & 0.189 & 0.231 & 0.132 & 0.159 & 0.314 & \cellcolor{macroHL} 0.205 & 0.224 & 0.296 & 0.252 & \cellcolor{macroHL} 0.257 \\
\midrule
\multicolumn{11}{l}{\emph{OLMo-Hybrid-7B-Instruct (GDN)}} \\
Full recompute & 0.568 & 0.468 & 0.312 & 0.219 & 0.394 & \cellcolor{macroHL} 0.392 & 0.213 & 0.292 & 0.227 & \cellcolor{macroHL} 0.244 \\
Na\"ive reuse & 0.306 & 0.206 & 0.109 & 0.086 & 0.237 & \cellcolor{macroHL} 0.189 & 0.206 & 0.285 & 0.224 & \cellcolor{macroHL} 0.238 \\
CacheBlend + \textsc{ex} & 0.244 & 0.171 & 0.119 & 0.083 & 0.191 & \cellcolor{macroHL} 0.162 & 0.203 & 0.298 & 0.220 & \cellcolor{macroHL} 0.240 \\
\rowcolor{linearKVGray} CacheBlend + \textsc{lb} & 0.227 & 0.174 & 0.110 & 0.086 & 0.186 & \cellcolor{macroHL} 0.157 & 0.212 & 0.296 & 0.217 & \cellcolor{macroHL} 0.242 \\
EPIC + \textsc{ex} & 0.469 & 0.346 & 0.246 & 0.162 & 0.350 & \cellcolor{macroHL} 0.315 & 0.207 & 0.309 & 0.229 & \cellcolor{macroHL} 0.248 \\
\rowcolor{linearKVGray} EPIC + \textsc{lb} & 0.459 & 0.337 & 0.240 & 0.164 & 0.352 & \cellcolor{macroHL} 0.310 & 0.203 & 0.307 & 0.227 & \cellcolor{macroHL} 0.246 \\
ProphetKV + \textsc{ex} & 0.510 & 0.413 & 0.278 & 0.182 & 0.359 & \cellcolor{macroHL} 0.348 & 0.199 & 0.287 & 0.153 & \cellcolor{macroHL} 0.213 \\
\rowcolor{linearKVGray} ProphetKV + \textsc{lb} & 0.518 & 0.421 & 0.306 & 0.179 & 0.380 & \cellcolor{macroHL} 0.361 & 0.198 & 0.285 & 0.163 & \cellcolor{macroHL} 0.216 \\
\midrule
\multicolumn{11}{l}{\emph{Qwen3.6-27B (GDN)}} \\
Full recompute$^{\ddagger}$ & 0.683 & 0.682 & 0.553 & 0.331 & 0.505 & \cellcolor{macroHL} 0.551 & 0.242 & 0.331 & 0.249 & \cellcolor{macroHL} 0.274 \\
Na\"ive reuse$^{\ddagger}$ & 0.458 & 0.423 & 0.232 & 0.182 & 0.414 & \cellcolor{macroHL} 0.342 & 0.204 & 0.307 & 0.239 & \cellcolor{macroHL} 0.250 \\
CacheBlend + \textsc{ex} & 0.545 & 0.496 & 0.388 & 0.252 & 0.463 & \cellcolor{macroHL} 0.429 & 0.211 & 0.327 & 0.245 & \cellcolor{macroHL} 0.261 \\
\rowcolor{linearKVGray} CacheBlend + \textsc{lb} & 0.540 & 0.504 & 0.361 & 0.259 & 0.459 & \cellcolor{macroHL} 0.425 & 0.212 & 0.323 & 0.243 & \cellcolor{macroHL} 0.259 \\
EPIC + \textsc{ex} & 0.624 & 0.544 & 0.420 & 0.286 & 0.488 & \cellcolor{macroHL} 0.473 & 0.220 & 0.326 & 0.248 & \cellcolor{macroHL} 0.265 \\
\rowcolor{linearKVGray} EPIC + \textsc{lb} & 0.615 & 0.539 & 0.410 & 0.290 & 0.484 & \cellcolor{macroHL} 0.468 & 0.220 & 0.324 & 0.249 & \cellcolor{macroHL} 0.264 \\
ProphetKV + \textsc{ex} & 0.620 & 0.498 & 0.408 & 0.273 & 0.397 & \cellcolor{macroHL} 0.439 & 0.221 & 0.327 & 0.244 & \cellcolor{macroHL} 0.264 \\
\rowcolor{linearKVGray} ProphetKV + \textsc{lb} & 0.613 & 0.477 & 0.396 & 0.258 & 0.388 & \cellcolor{macroHL} 0.426 & 0.228 & 0.323 & 0.248 & \cellcolor{macroHL} 0.266 \\
\bottomrule
\end{tabular}
\end{table}

\begin{table}[t]
\centering
\caption{RULER results (string-match recall) under the same matched ${\sim}20\%$ budget ($r{=}0.2$): per subtask at 8K and at 32K, each with its Avg highlighted. Row conventions follow Table~\ref{tab:main}. Only four subtasks are run at 32K, so the two Avg columns are over different subtask sets and are not directly comparable. Subtasks: common-/frequent-word extraction (CWE, FWE), single/multi-key/multi-query/multi-value needle retrieval (N-S, N-MK, N-MQ, N-MV), SQuAD-style QA, variable tracking (VT). Granite's full-recompute CWE cell ($0.003$, below its own na\"ive-reuse score) is a model-level failure on that subtask rather than a reuse or scoring artifact.}
\label{tab:ruler}
\scriptsize
\setlength{\tabcolsep}{3pt}
\begin{tabular}{lcccccccc>{\columncolor{macroHL}}ccccc>{\columncolor{macroHL}}c}
\toprule
& \multicolumn{9}{c}{RULER-8K, 8 subtasks} & \multicolumn{5}{c}{RULER-32K, 4 subtasks} \\
\cmidrule(lr){2-10} \cmidrule(lr){11-15}
Method & CWE & FWE & N-S & N-MK & N-MQ & N-MV & QA & VT & Avg & N-S & N-MK & VT & QA & Avg \\
\midrule
\multicolumn{15}{l}{\emph{Granite-4.0-H-Tiny (Mamba-2)}} \\
Full recompute & 0.003 & 0.963 & 1.000 & 0.930 & 0.960 & 0.760 & 0.670 & 0.314 & \cellcolor{macroHL} 0.700 & 1.000 & 0.650 & 0.498 & 0.340 & \cellcolor{macroHL} 0.622 \\
Na\"ive reuse & 0.192 & 0.897 & 0.110 & 0.230 & 0.263 & 0.212 & 0.240 & 0.134 & \cellcolor{macroHL} 0.285 & 0.010 & 0.010 & 0.018 & 0.120 & \cellcolor{macroHL} 0.040 \\
CacheBlend + \textsc{ex} & 0.002 & 0.020 & 0.000 & 0.000 & 0.000 & 0.000 & 0.220 & 0.000 & \cellcolor{macroHL} 0.030 & 0.000 & 0.000 & 0.000 & 0.170 & \cellcolor{macroHL} 0.043 \\
\rowcolor{linearKVGray} CacheBlend + \textsc{lb} & 0.052 & 0.547 & 0.970 & 0.220 & 0.297 & 0.318 & 0.460 & 0.140 & \cellcolor{macroHL} 0.375 & 0.960 & 0.270 & 0.036 & 0.410 & \cellcolor{macroHL} 0.419 \\
EPIC + \textsc{ex} & 0.040 & 0.183 & 0.030 & 0.070 & 0.100 & 0.090 & 0.320 & 0.044 & \cellcolor{macroHL} 0.110 & 0.010 & 0.000 & 0.004 & 0.260 & \cellcolor{macroHL} 0.069 \\
\rowcolor{linearKVGray} EPIC + \textsc{lb} & 0.129 & 0.913 & 0.900 & 0.670 & 0.820 & 0.542 & 0.550 & 0.240 & \cellcolor{macroHL} 0.596 & 1.000 & 0.730 & 0.214 & 0.580 & \cellcolor{macroHL} 0.631 \\
ProphetKV + \textsc{ex} & 0.021 & 0.133 & 0.930 & 0.370 & 0.510 & 0.320 & 0.370 & 0.238 & \cellcolor{macroHL} 0.362 & 0.030 & 0.000 & 0.030 & 0.250 & \cellcolor{macroHL} 0.077 \\
\rowcolor{linearKVGray} ProphetKV + \textsc{lb} & 0.027 & 0.810 & 1.000 & 0.720 & 0.848 & 0.745 & 0.510 & 0.320 & \cellcolor{macroHL} 0.622 & 1.000 & 0.520 & 0.306 & 0.600 & \cellcolor{macroHL} 0.607 \\
\midrule
\multicolumn{15}{l}{\emph{OLMo-Hybrid-7B-Instruct (GDN)}} \\
Full recompute & 0.869 & 0.850 & 1.000 & 0.990 & 0.990 & 0.983 & 0.750 & 0.394 & \cellcolor{macroHL} 0.853 & 1.000 & 0.990 & 0.234 & 0.770 & \cellcolor{macroHL} 0.749 \\
Na\"ive reuse & 0.316 & 0.903 & 0.810 & 0.540 & 0.260 & 0.168 & 0.410 & 0.302 & \cellcolor{macroHL} 0.464 & 0.020 & 0.140 & 0.074 & 0.210 & \cellcolor{macroHL} 0.111 \\
CacheBlend + \textsc{ex} & 0.374 & 0.937 & 0.980 & 0.420 & 0.302 & 0.175 & 0.490 & 0.122 & \cellcolor{macroHL} 0.475 & 0.760 & 0.190 & 0.254 & 0.330 & \cellcolor{macroHL} 0.384 \\
\rowcolor{linearKVGray} CacheBlend + \textsc{lb} & 0.331 & 0.933 & 0.980 & 0.430 & 0.300 & 0.172 & 0.460 & 0.140 & \cellcolor{macroHL} 0.468 & 0.750 & 0.160 & 0.274 & 0.290 & \cellcolor{macroHL} 0.368 \\
EPIC + \textsc{ex} & 0.537 & 0.913 & 1.000 & 0.700 & 0.767 & 0.388 & 0.660 & 0.232 & \cellcolor{macroHL} 0.650 & 1.000 & 0.590 & 0.252 & 0.470 & \cellcolor{macroHL} 0.578 \\
\rowcolor{linearKVGray} EPIC + \textsc{lb} & 0.533 & 0.900 & 1.000 & 0.710 & 0.760 & 0.378 & 0.660 & 0.266 & \cellcolor{macroHL} 0.651 & 1.000 & 0.610 & 0.272 & 0.460 & \cellcolor{macroHL} 0.586 \\
ProphetKV + \textsc{ex} & 0.616 & 0.873 & 1.000 & 0.940 & 0.970 & 0.940 & 0.580 & 0.382 & \cellcolor{macroHL} 0.788 & 1.000 & 0.730 & 0.428 & 0.540 & \cellcolor{macroHL} 0.674 \\
\rowcolor{linearKVGray} ProphetKV + \textsc{lb} & 0.600 & 0.877 & 1.000 & 0.900 & 0.968 & 0.950 & 0.580 & 0.430 & \cellcolor{macroHL} 0.788 & 1.000 & 0.750 & 0.420 & 0.540 & \cellcolor{macroHL} 0.677 \\
\midrule
\multicolumn{15}{l}{\emph{Qwen3.6-27B (GDN)}} \\
Full recompute & 0.998 & 0.997 & 1.000 & 1.000 & 1.000 & 1.000 & 0.870 & 1.000 & \cellcolor{macroHL} 0.983 & 1.000 & 1.000 & 1.000 & 0.920 & \cellcolor{macroHL} 0.980 \\
Na\"ive reuse & 0.079 & 0.993 & 0.990 & 0.470 & 0.370 & 0.302 & 0.710 & 0.100 & \cellcolor{macroHL} 0.502 & 0.950 & 0.390 & 0.204 & 0.580 & \cellcolor{macroHL} 0.531 \\
CacheBlend + \textsc{ex} & 0.807 & 0.993 & 0.980 & 0.660 & 0.708 & 0.318 & 0.670 & 0.294 & \cellcolor{macroHL} 0.679 & 0.980 & 0.380 & 0.218 & 0.700 & \cellcolor{macroHL} 0.570 \\
\rowcolor{linearKVGray} CacheBlend + \textsc{lb} & 0.811 & 0.993 & 0.980 & 0.660 & 0.688 & 0.320 & 0.680 & 0.266 & \cellcolor{macroHL} 0.675 & 0.980 & 0.370 & 0.226 & 0.690 & \cellcolor{macroHL} 0.567 \\
EPIC + \textsc{ex} & 0.945 & 1.000 & 0.970 & 0.680 & 0.782 & 0.403 & 0.830 & 0.306 & \cellcolor{macroHL} 0.740 & 0.990 & 0.610 & 0.270 & 0.830 & \cellcolor{macroHL} 0.675 \\
\rowcolor{linearKVGray} EPIC + \textsc{lb} & 0.961 & 1.000 & 0.970 & 0.680 & 0.765 & 0.415 & 0.830 & 0.280 & \cellcolor{macroHL} 0.738 & 0.990 & 0.610 & 0.268 & 0.840 & \cellcolor{macroHL} 0.677 \\
ProphetKV + \textsc{ex} & 0.719 & 0.983 & 0.990 & 0.860 & 0.873 & 0.660 & 0.580 & 0.918 & \cellcolor{macroHL} 0.823 & 0.950 & 0.690 & 0.978 & 0.700 & \cellcolor{macroHL} 0.829 \\
\rowcolor{linearKVGray} ProphetKV + \textsc{lb} & 0.696 & 0.977 & 0.990 & 0.880 & 0.885 & 0.645 & 0.600 & 0.922 & \cellcolor{macroHL} 0.824 & 0.950 & 0.690 & 0.978 & 0.700 & \cellcolor{macroHL} 0.829 \\
\bottomrule
\end{tabular}
\end{table}

\subsection{Setup}

\paragraph{Models.} We evaluate three hybrid LLMs spanning both recurrence families, two scales, and different linear-to-FA ratios. \textbf{Granite-4.0-H-Tiny}~\citep{graniteteam2025granite} ($7$B, MoE) has \emph{Mamba-2} (scalar-decay) linear layers---$4$ FA and $36$ Mamba-2 of $40$ layers ($1{:}9$, $90\%$ recurrent). \textbf{OLMo-Hybrid-7B-Instruct}~\citep{merrill2026olmohybrid} ($7$B) and \textbf{Qwen3.6-27B}~\citep{qwen2026qwen36} ($27$B) use \emph{Gated DeltaNet} (GDN; dense transition): $8$ FA/$24$ GDN of $32$ layers and $16$ FA/$48$ GDN of $64$ layers, respectively (both $1{:}3$, $75\%$ recurrent).

\paragraph{Data and metrics.} Following prior PIC work \citep{yao2025cacheblend,hu2024epic,prophetkv2026}, we use long-context benchmarks whose prompts pair a large reusable context with a short query (the retrieval-augmented generation, RAG, setting) \citep{lewis2020rag}: (i)~five LongBench QA datasets \citep{bai2024longbench}---HotpotQA (HQA), 2WikiMQA (2Wiki), MuSiQue (MSQ), NarrativeQA (NQA), Qasper (Qsp); (ii)~RULER \citep{hsieh2024ruler} at 8K--32K, with subtasks common-/frequent-word extraction (CWE, FWE), single/multi-key/multi-query/multi-value needle retrieval (N-S, N-MK, N-MQ, N-MV), SQuAD-style QA, and variable tracking (VT); and (iii)~three LongBench summarization datasets (QMSum, GovReport, MultiNews; 512-token generation). LongBench QA uses $n{=}200$; RULER (both 8K and 32K) and LongBench summarization use $n{=}100$. Quality is token-level F1 for QA (averaged as Avg-F1), string-match recall for RULER, and ROUGE-L \citep{lin2004rouge} for summarization; we also report \%-of-full $=$ score$/$full-recompute$\times100$. Following ProphetKV, each context is split into fixed, contiguous 512-token chunks, each prefilled and cached independently. Decoding is greedy and every selector is deterministic, so runs are reproducible up to the recurrent \texttt{fla}/Mamba-2 Triton kernels' nondeterminism, which leaves a small run-to-run spread on Mamba-2; we therefore read the two GDN initializers (paired \textsc{ex}/\textsc{lb} gap $\le0.013$ Avg-F1 and sign-changing) as interchangeable, while the Granite gaps we highlight (e.g.\ $0.145$ vs.\ $0.270$ under EPIC) are an order of magnitude larger.

\paragraph{Methods.} The selector $\Pi$ is CacheBlend, EPIC, or ProphetKV; we pair exact composition and LinearKV under the same selected positions and recomputation. Two anchors bound every table: \emph{full recompute} ($r{=}1$, lossless upper bound) and \emph{na\"ive reuse} ($r{=}0$, the last-block state with no recomputation, lower bound). The \emph{effective recompute ratio} is $r=|\mathcal{U}|/N$, the fraction of the $N$ context tokens the selector recomputes; since both initializers process the identical $\mathcal{U}$ through identical layers, any quality difference is attributable to $S_{\mathrm{init}}$ alone. Main-table runs use matched $r{\approx}0.2$; sweeps cover $r\in\{0.03,0.05,0.1,0.2,0.3,0.4\}$.

\paragraph{Environment.} Experiments run on NVIDIA H800 GPUs (80\,GB) using HuggingFace Transformers \citep{wolf2020transformers} with \texttt{flash-linear-attention} kernels \citep{yang2024fla} in \texttt{bfloat16}; each model fits one device, so GPUs shard the workload across datasets. Cached artifacts---per-chunk FA KV and each recurrent layer's end state (plus, for exact composition, the per-token key/gate stream folded online)---are stored in \texttt{bfloat16} and loaded from host memory at serving time. TTFT (Table~\ref{tab:ttft}) is measured on a single exclusive GPU.

\subsection{Main Results}

Tables~\ref{tab:main} and~\ref{tab:ruler} evaluate LinearKV as a hybrid-PIC framework: three existing PIC selectors---CacheBlend, EPIC, and ProphetKV---plug in unchanged, each run with two initializers under identical selected positions and recomputation---exact composition (\textsc{ex}) and LinearKV's last-block state (\textsc{lb})---at a matched ${\sim}20\%$ budget, with full recompute and na\"ive reuse as anchors. All four benchmarks are reported per dataset or subtask: token-F1 on LongBench QA and ROUGE-L on LongBench summarization (Table~\ref{tab:main}), and string-match recall on RULER at 8K and 32K (Table~\ref{tab:ruler}). Summarization and RULER-32K confirm the split holds under generation and at longer context.

Our first finding concerns the framework itself: with nothing more than the decoupled initialization, every existing selector transfers to hybrid models. On the two GDN models the choice of linear-state initializer is immaterial---the paired \textsc{ex}/\textsc{lb} gap is at most $0.013$ Avg-F1 and even changes sign across selectors---yet both recover most of full quality, reaching $92\%$ under ProphetKV (OLMo, RULER-8K Avg; Table~\ref{tab:ruler}).

The second finding is more surprising: the equivalence breaks on Mamba-2, and in the counterintuitive direction the principled construction is the fragile one. On Granite, \textsc{ex} collapses under \emph{every} selector while \textsc{lb} stays robust (Avg-F1 $0.063/0.145/0.125$ vs.\ $0.118/0.270/0.205$ for CacheBlend/EPIC/ProphetKV). The gap is largest under EPIC, where changing \emph{only} the initializer---not the selector or the recomputed positions---lifts quality from $46.6\%$ to $86.8\%$ of full. RULER reproduces the split under a recall metric (Table~\ref{tab:ruler}; Granite Avg $0.030/0.110/0.362$ vs.\ $0.375/0.596/0.622$), and it persists at $4\times$ context (RULER-32K Avg, Granite $0.043/0.069/0.077$ vs.\ $0.419/0.631/0.607$) and under generation on summarization (Table~\ref{tab:main}, Summ Avg, Granite $0.146/0.197/0.198$ vs.\ $0.245/0.269/0.257$).\footnote{One cell warrants explicit comment: Granite's full-recompute score on RULER CWE is $0.003$, \emph{below} its own na\"ive-reuse score of $0.192$. This is a model-level failure on that subtask, not a reuse artifact or a scoring bug: under the identical pipeline, prompt, and metric, OLMo and Qwen reach $0.869$ and $0.998$, and Granite's answers there are well formed but list words with no overlap with the reference set. Excluding CWE, Granite's RULER-8K full-recompute average is $0.800$ rather than $0.700$, and the \%-of-full figures we report for Granite on RULER shift by at most $2.3$pp; we retain CWE in the average rather than dropping an unfavorable subtask.} Because last-block matches exact composition on GDN yet dominates it on Mamba-2 under every selector, LinearKV adopts it as the default; the ablations below trace the collapse to the multi-source construction, not the architecture.

\subsection{Efficiency: TTFT comparison}

\begin{table}[t]
\centering
\caption{TTFT across models and context lengths at a matched recomputation ratio $r{=}0.2$, with the initial-state construction held side by side: \textsc{lb} = LinearKV and \textsc{ex} = exact composition. Each cell is TTFT in milliseconds (median over $4$ documents $\times$ $3$ repeats; $p_{90}$ within $2\%$).}
\label{tab:ttft}
\footnotesize
\setlength{\tabcolsep}{4pt}
\begin{tabular}{l|c|cc|cc|cc|cc}
\toprule
 & & & & \multicolumn{2}{c|}{CacheBlend} & \multicolumn{2}{c|}{EPIC} & \multicolumn{2}{c}{ProphetKV} \\
\cmidrule(lr){5-6} \cmidrule(lr){7-8} \cmidrule(lr){9-10}
Model & Ctx & Full & Na\"ive & \textsc{lb} & \textsc{ex} & \textsc{lb} & \textsc{ex} & \textsc{lb} & \textsc{ex} \\
\midrule
\multirow{3}{*}{Granite (Mamba-2)} & 32K & 782 & 112 & \textbf{397} & 438 & \textbf{377} & 429 & \textbf{482} & 543 \\
 & 16K & 429 & 102 & \textbf{261} & 291 & \textbf{258} & 287 & \textbf{367} & 399 \\
 & 8K & 329 & 117 & \textbf{264} & 283 & \textbf{266} & 284 & \textbf{296} & 314 \\
\midrule
\multirow{3}{*}{OLMo (GDN)} & 32K & 1129 & 44 & \textbf{520} & 593 & \textbf{518} & 595 & \textbf{571} & 649 \\
 & 16K & 524 & 41 & \textbf{220} & 253 & \textbf{217} & 253 & \textbf{266} & 300 \\
 & 8K & 248 & 43 & \textbf{116} & 136 & \textbf{116} & 134 & \textbf{161} & 179 \\
\midrule
\multirow{3}{*}{Qwen (GDN)} & 32K & 3843 & 90 & \textbf{1810} & 1967 & \textbf{1827} & 1984 & \textbf{1955} & 2109 \\
 & 16K & 1750 & 81 & \textbf{666} & 750 & \textbf{663} & 747 & \textbf{775} & 859 \\
 & 8K & 841 & 90 & \textbf{342} & 396 & \textbf{343} & 395 & \textbf{419} & 457 \\
\bottomrule
\end{tabular}
\end{table}

Having shown that the last-block initial state matches or exceeds exact composition on quality, we turn to cost. Table~\ref{tab:ttft} reports TTFT at $r{=}0.2$. Both initializers make reuse a clear win over full prefix recomputation: in every model, context, and selector, TTFT stays well below a full recompute---at 32K last-block needs only $0.46$--$0.62\times$ full-recomputation TTFT and exact composition ${\le}0.69\times$, and the benefit already holds at 8K. Hybrid PIC therefore pays off regardless of how the linear state is initialized.

Between the two, last-block is uniformly cheaper: it is faster than exact composition in \emph{all $27$ measured pairs}, by $5$--$17\%$: on GDN, exact composition must online-fold a dense per-chunk transition $T_{C_j}$, whereas last-block reads a single state and folds nothing. Last-block thus wins on \emph{both} axes---the better end-to-end quality of Table~\ref{tab:main} \emph{and} the lower TTFT here---further confirming its advantage. Efficiency is single-request TTFT on an idle GPU; batched serving and host-to-device cache transfer are future work.

\subsection{Ablation Studies}

Having shown that the last-block initial state wins on both quality (Table~\ref{tab:main}) and latency (Table~\ref{tab:ttft}), we now examine exact composition's failure: more recompute does not fix it, and it tracks the multi-source construction rather than the architecture.

\begin{figure}[t]
\centering
\includegraphics[width=0.56\textwidth]{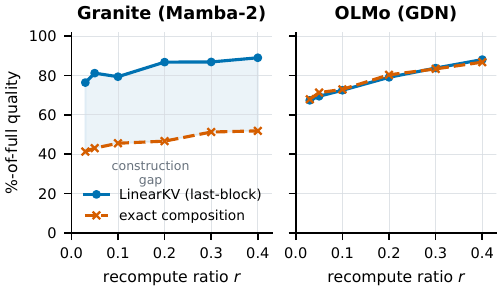}
\caption{Exact composition init vs.\ lastblock init across recompute ratios (EPIC selector; Granite, OLMo).}
\label{fig:ratiosweep}
\end{figure}

\paragraph{A bad initializer cannot be rescued by more recompute.} We hold the selector, its positions, and the recomputation fixed, and vary \emph{only} the linear-state initializer---exact composition versus last-block---across recompute budgets $r\in\{0.03,0.05,0.1,0.2,0.3,0.4\}$ (Figure~\ref{fig:ratiosweep}, EPIC selector). On Granite (Mamba-2), exact composition stays flat at $41$--$52\%$ of full quality at \emph{every} budget while last-block reaches $76$--$89\%$; the $35$--$40$pp gap never closes, even at $r{=}0.4$. The initializer therefore sets a quality ceiling that added recompute cannot lift---a bad initial state is not bought back with more repair. This also locates the failure: because last-block, a \emph{different} initializer, recovers most of full quality on the \emph{same} model, exact's collapse is a property of its multi-source construction, not a Mamba-2 reuse ceiling (on OLMo the two coincide at every budget). The same gap holds across selectors at $r{=}0.2$: on Granite, exact composition loses $0.055$--$0.125$ Avg-F1 to the last-block initial state under CacheBlend, EPIC, and ProphetKV---largest under EPIC, whose cell drops from $0.270$ to $0.145$, i.e.\ from $86.8\%$ to $46.6\%$ of full quality---while on the two GDN models the paired gap stays within $\pm0.013$ and changes sign. This asymmetry matches the depth-wise error diagnosis of Figure~\ref{fig:prerelerr}: on Mamba-2 the composition error compounds through depth, while on GDN it stays bounded.

\begin{figure}[t]
\centering
\includegraphics[width=0.72\textwidth]{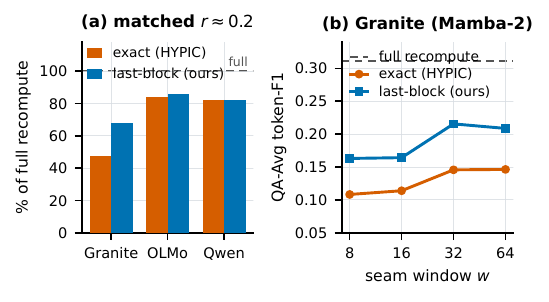}
\caption{End-to-end comparison of LinearKV and HYPIC. For fairness both use HYPIC's seam selector; the only variable is the initializer---HYPIC's exact composition vs.\ LinearKV's lastblock init.}
\label{fig:hypic}
\end{figure}

\paragraph{Comparison with a HYPIC-style seam selector.} Concurrent HYPIC pairs a \emph{seam} selector---recomputing a fixed window around each chunk boundary---with exact composition. We compare LinearKV against HYPIC's initializer under this shared seam selector, holding the seam positions fixed and changing only the linear-state initializer---HYPIC's exact composition vs.\ LinearKV's last block (Figure~\ref{fig:hypic}); to keep the initializer the sole variable, neither side adds HYPIC's remaining pipeline stages (boundary-token exclusion, per-seam recomposition, causal-convolution warm-up). The main-table pattern survives intact. At a matched ${\sim}20\%$ budget (Figure~\ref{fig:hypic}a), replacing exact composition with the last-block initializer lifts Granite from $47.6\%$ to $67.8\%$ of full recompute---a $20$pp gain from the initializer alone---while OLMo and Qwen stay within $2$pp. Sweeping the seam window on Granite (Figure~\ref{fig:hypic}b) shows the same separation at every budget, with no sign of closing. LinearKV's last-block initializer therefore dominates HYPIC's exact composition under the shared seam selector: it removes the Mamba-2 fragility and costs nothing on GDN.

\section{Related Work}
\label{sec:related}

\paragraph{Position-independent caching for LLMs.} PIC reuses independently prefilled chunks regardless of prefix, recomputing a few tokens to restore cross-chunk context. For full-attention models, CacheBlend \citep{yao2025cacheblend}, EPIC \citep{hu2024epic}, KVShare \citep{yang2025kvshare}, ProphetKV \citep{prophetkv2026}, KVLink \citep{yang2025kvlink}, PromptCache \citep{gim2024promptcache}, CacheClip \citep{yang2025cacheclip}, Cache-Craft \citep{agarwal2025cachecraft}, CacheSlide \citep{liu2026cacheslide}, and MiniPIC \citep{ordonez2026minipic} vary the chunking, token-selection, repair, and engine-integration policies, while MEPIC \citep{wang2025mepic} and C$^2$KV \citep{du2026c2kv} make the reused KV memory-efficient and composable. A parallel line adapts PIC to non-token-indexed representations: Irminsul \citep{ma2026irminsul} for multi-head latent attention, COMB \citep{zhao2026encoder} via a retrained encoder, and concurrent HYPIC \citep{liu2026hypic} for hybrid LLMs via exact operator composition. LinearKV joins this line for the recurrent-state representation, training-free: it keeps the compile/link workflow and existing selectors unchanged, and its contribution begins where the token-indexed abstraction ends---how the same selected positions initialize and advance recurrent states to jointly repair a hybrid cache.

\paragraph{SSM and hybrid-state caching.} Marconi \citep{pan2025marconi}, sparse prefix caching \citep{shirokikh2026sparse}, and compiler-level SSD caching \citep{santoni2026compiler} reuse recurrent states only at exact prefix checkpoints, not under lossy assembly of independently prefilled segments. Recurrent-state error has been studied from an error-control view \citep{chung2026statetracking}; we identify independent-prefill hidden-input mismatch as a concrete source.

\paragraph{AI-infrastructure positioning.} Cache reuse instantiates the computation--memory leg of the computation--bandwidth--memory trade-off \citep{fan2026cbm}; the same lens motivates generative, task-oriented transmission when bandwidth binds \citep{chen2026generative}. Hybrid LLMs are natural device--edge targets \citep{an2026aiflow}, where a principled quality--recompute dial beats an all-or-nothing cache.

\section{Conclusion}
\label{sec:conclusion}

We studied cross-request cache reuse for hybrid LLMs and showed that composing all $K$ cached states into the exact full-prefix state is input-conditional and architecture-fragile---harmful on the evaluated Mamba-2 model and unnecessary on two GDN models---because independently prefilled deep operators are built from context-mismatched hidden inputs. LinearKV instead keeps a single cached block as the initial state and turns any existing selector's positions into an ordered recomputation stream that repairs recurrent states and FA KV jointly, raising Mamba-2 quality from $46.6\%$ to $86.8\%$ of full under a matched budget while avoiding exact composition's $5$--$17\%$ online overhead. The result holds across LongBench QA and RULER at 8K--32K.

\bibliography{linearkv}
\bibliographystyle{iclr2027_conference}

\appendix
\clearpage
\section{Single-source ablation: last block vs.\ random block}
\label{app:singlesource}

A natural worry is that our result hinges on the last block in particular---its recency, or its adjacency to the query---rather than on using a single source. It does not. We add a \emph{random-block} control that initializes each recurrent layer from one \emph{randomly chosen} matched chunk's cached state (seeded; same selector, positions, and budget), and compare all three initializers at $r{=}0.2$ (Table~\ref{tab:singlesource}). On Granite (Mamba-2) the random block tracks the last block---within $0.014$ Avg-F1 at every selector---and both roughly double exact composition, so what matters is reading from a \emph{single} cached source, not which source. On the two GDN models all three initializers agree within $0.02$ Avg-F1, as expected where composition does not compound.

Two further controls delimit the effect: na\"ive reuse (Table~\ref{tab:main})---last-block \emph{without} recomputation---trails every recomputed single-source initializer, so recomputation is necessary; and the three selectors, choosing different positions, all clear na\"ive reuse, so the selected positions matter too. Last-block is therefore not a special trick but the natural zero-cost instance of single-source initialization---the most recent chunk, already cached, needing no extra selection---which is why we adopt it as the default.

\begin{table}[h]
\centering
\caption{Single-block init (random vs.\ last-block) compared with exact composition. Granite (Mamba-2), Avg-F1 at $r{=}0.2$.}
\label{tab:singlesource}
\small
\setlength{\tabcolsep}{8pt}
\begin{tabular}{lccc}
\toprule
Granite (Mamba-2) & exact & last-block & random-block \\
Avg-F1 @ $r{=}0.2$ & (multi-source) & (single) & (single) \\
\midrule
CacheBlend & 0.063 & 0.118 & 0.129 \\
EPIC & 0.145 & 0.270 & 0.256 \\
ProphetKV & 0.125 & 0.205 & 0.214 \\
\bottomrule
\end{tabular}
\end{table}

\end{document}